\documentclass[sn-mathphys-num]{sn-jnl}

\usepackage[numbers]{natbib}
\usepackage{graphicx}%
\usepackage{multirow}%
\usepackage{amsmath,amssymb,amsfonts}%
\usepackage{amsthm}%
\usepackage{mathrsfs}%
\usepackage[title]{appendix}%
\usepackage{xcolor}%
\usepackage{textcomp}%
\usepackage{manyfoot}%
\usepackage{booktabs}%
\usepackage{algorithm}%
\usepackage{algorithmicx}%
\usepackage{algpseudocode}%
\usepackage{listings}%
\usepackage{pgfplots}
\usepackage{tikz}
\usepackage{lineno}

\theoremstyle{thmstyleone}%
\theoremstyle{thmstyletwo}%

\theoremstyle{thmstylethree}%

\definecolor{darkblue1}{RGB}{65, 84, 119}
\definecolor{darkpurple1}{RGB}{165, 110, 225}
\definecolor{darkpink1}{RGB}{238, 83, 150}
\definecolor{darkteal1}{RGB}{0, 157, 154}

\begin{document}

\title[Coalitions of Large Language Models Increase the Robustness of AI Agents]{Coalitions of Large Language Models Increase the Robustness of AI Agents}


\author*[1]{\fnm{Prattyush} \sur{Mangal}}\email{prattyush.mangal@ibm.com}

\author[1]{\fnm{Carol} \sur{Mak}}
\author[1]{\fnm{Theo} \sur{Kanakis}}
\author[1]{\fnm{Timothy} \sur{Donovan}}
\author[1]{\fnm{Dave} \sur{Braines}}
\author*[1]{\fnm{Edward} \sur{Pyzer-Knapp}}\email{epyzerk3@uk.ibm.com}

\affil[1]{\orgdiv{IBM Research}, \country{Europe (UK)}}


\abstract{
The emergence of Large Language Models (LLMs) have fundamentally altered the way we interact with digital systems and have led to the pursuit of LLM powered AI agents to assist in daily workflows. LLMs, whilst powerful and capable of demonstrating some emergent properties, are not logical reasoners and often struggle to perform well at all sub-tasks carried out by an AI agent to plan and execute a workflow. While existing studies tackle this lack of proficiency by generalised pretraining at a huge scale or by specialised fine-tuning for tool use, we assess if a system comprising of a coalition of pretrained LLMs, each exhibiting specialised performance at individual sub-tasks, can match the performance of single model agents. The coalition of models approach showcases its potential for building robustness and reducing the operational costs of these AI agents by leveraging traits exhibited by specific models. Our findings demonstrate that fine-tuning can be mitigated by considering a coalition of pretrained models and believe that this approach can be applied to other non-agentic systems which utilise LLMs.
}


\maketitle

\section{Introduction}\label{sec1}

\begin{figure*}
    \centering
    \includegraphics[width=1\linewidth]{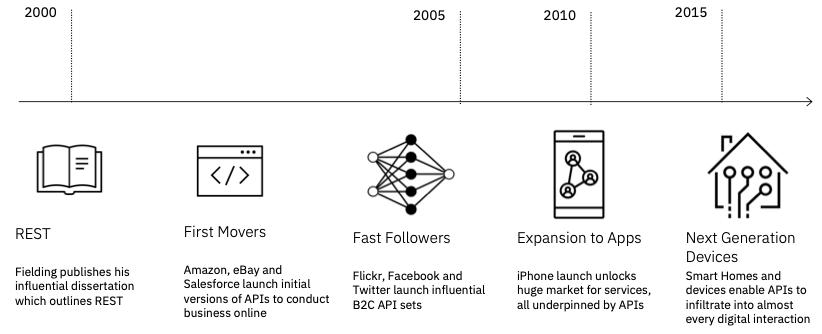}
    \caption{Timeline depicting major events in the utilisation of APIs.}
    \label{fig:api_utilisation}
\end{figure*}

Recent advances in AI methodologies, primarily driven by the emergence of Large Language Models (LLMs) \cite{zhao2023llmsurvey, devlin2018bert, touvron2023llama} have fundamentally altered the way we interact with digital systems \cite{wang2024agentssurvey, wu2023autogen}. This advance has coalesced with a longer, more deliberate set of advances in the way that machines communicate through the internet to request and consume the output of, services – namely the development of the API (Figure \ref{fig:api_utilisation}) \cite{fielding2000restapi, ofoeda2019apissurvey}. Given the ubiquity of their utilisation, it is therefore natural to consider whether LLMs are able to provide utility to the construction, execution and analysis of workflows constructed of API calls. 

LLMs, whilst powerful and capable of demonstrating some emergent properties, are not logical reasoners or planners \cite{kambhampati2024llmplanning}. However, we believe that through careful dissection and reasonably rich service description, there is a path to utility from which further advances can be built.  It should be noted that throughout this work we use the term ‘planning’ loosely, to refer to the construction of useful workflows, verifiable through comparison to a gold standard, not the creative task of optimized planning, or indeed reasoning.  

\begin{figure*}
    \centering
    \includegraphics[width=1\linewidth]{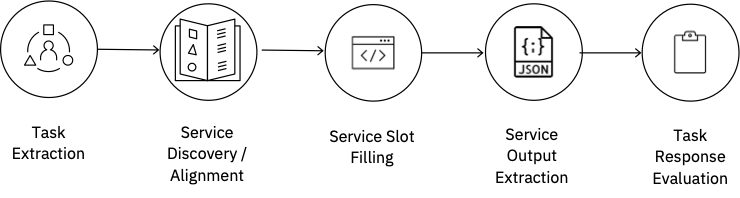}
    \caption{Stages in a decomposed agentic workflow for API consumption as evaluated in this study.}
    \label{fig:decomposed_workflow}
\end{figure*}

\begin{figure*}
    \centering
    \includegraphics[width=1\linewidth]{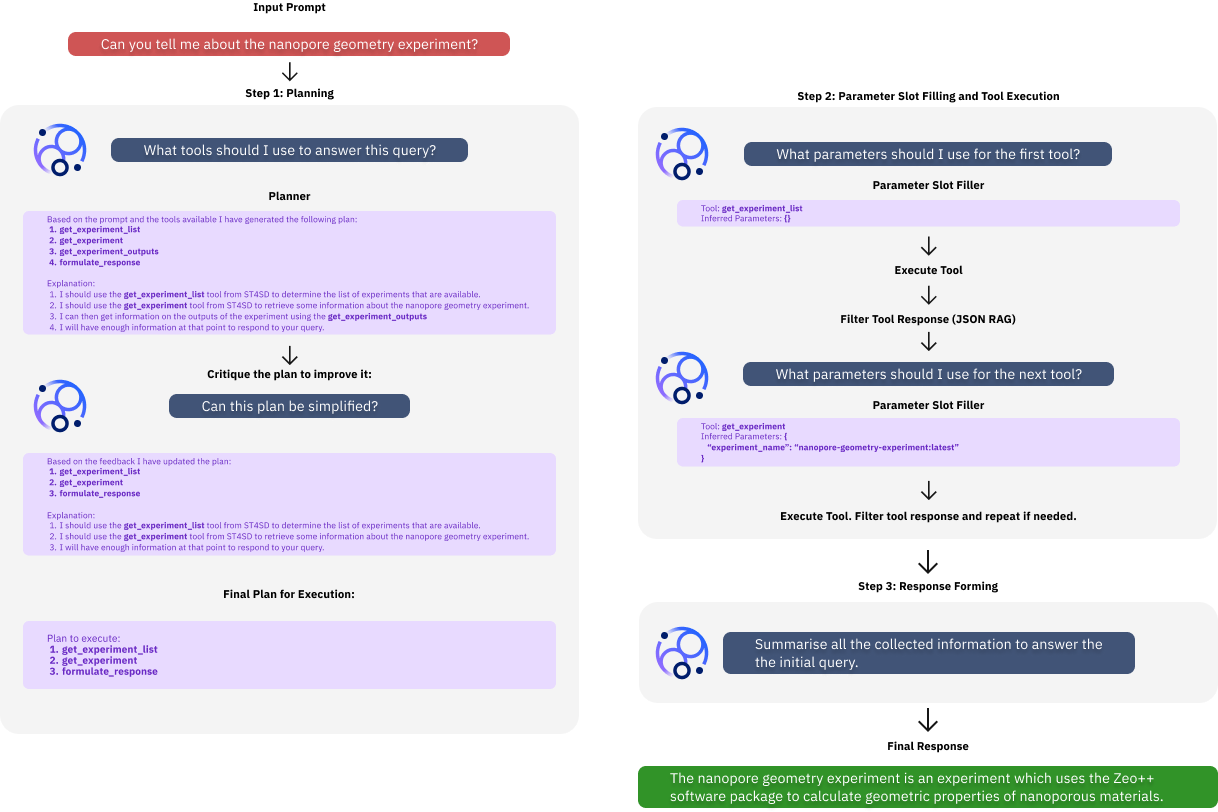}
    \caption{An overview of the workflow to answer intents and queries by orchestrating calls to external tools. Sub-tasks involve planning tool usage, slot filling tool parameters and summarising the collected information to form a final natural language response to the initial query. Each sub-task is assigned to different LLMs to achieve more accurate workflow executions. Example tasks allocated to LLMs are identified by blue prompts and LLM responses are identified by purple messages.}
    \label{fig:detailed_processing}
\end{figure*}

In this paper, we follow a decomposition approach to workflow construction, see Figure \ref{fig:decomposed_workflow}, in which tasks are first extracted from a natural language interaction, then compared to a catalogue of capabilities known to the agent.  After services are identified, we then task the system with constructing correct (both in terms of content and semantics) payloads for these services (referred to as slot filling), which are then executed. Finally, the output from these services is then extracted and summarised into a natural language form, ready for consumption by the human agent. See Figure \ref{fig:detailed_processing} for an example of the complete workflow.

In classical approaches the LLMs used have been singular pretrained models with a large amount of model parameters (GPT-like models) \cite{achiam2023openaigpt, song2023restgpt, OpenGPT}  or singular fine-tuned models specifically tuned for tool use tasks \cite{qin2023toolllm, tang2023toolalpaca, schick2024toolformer, patil2023gorilla}. The systems built around these models make use of different prompting techniques such as ReACT \cite{yao2022react}, Chain of Thought \cite{wei2022chainofthought}, Plan-and-solve \cite{planandsolve} although others have been developed \cite{critiqueprompting, reflexion}. Whilst methodologies based around these approaches have had some successes they are subject to several limitations:

\begin{enumerate}
    \item Specificity:  Single model methods require a sufficiently general model to work across tasks, which requires either significant training or task specific tuning; both of which incur significant cost.
    \item Disruption: Single model methods are prone to disruption by the development of newer models, which might require a different approach to prompt construction.
    \item Cost: Models which are sufficiently general are likely to have large parameter counts, which are typically associated with large deployment requirements, and therefore costs.
\end{enumerate}

We propose a system which allocates the tasks of service identification and discovery, slot filling, and response forming to different open-source models, to increase robustness and address these limitations.  Through demonstration on the ToolAlpaca benchmark \citep{tang2023toolalpaca} we show that this coalition approach improves upon standards, reached through solely using commonly used base models. We also show that, surprisingly, this coalition approach can outperform a single model fine-tuning approach and hence avoids incurring any of the costs of fine-tuning – although we note that this result is likely not general and fine tuning models specifically for model-task pairs would allow for more transferable performance gains across a broader workflow composition spectrum.

\section{Results}

The most common approaches for building AI agent systems for processing complex workflows fit into two categories - pretraining models at a huge scale or relying on fine-tuning models to unlock tool use capabilities \cite{achiam2023openaigpt, qin2023toolllm, tang2023toolalpaca, schick2024toolformer, patil2023gorilla}. In both cases, there are associated costs of data generation and compute resources to learn the new model parameters. 

\begin{figure}
    \centering
    \includegraphics[width=1\linewidth]{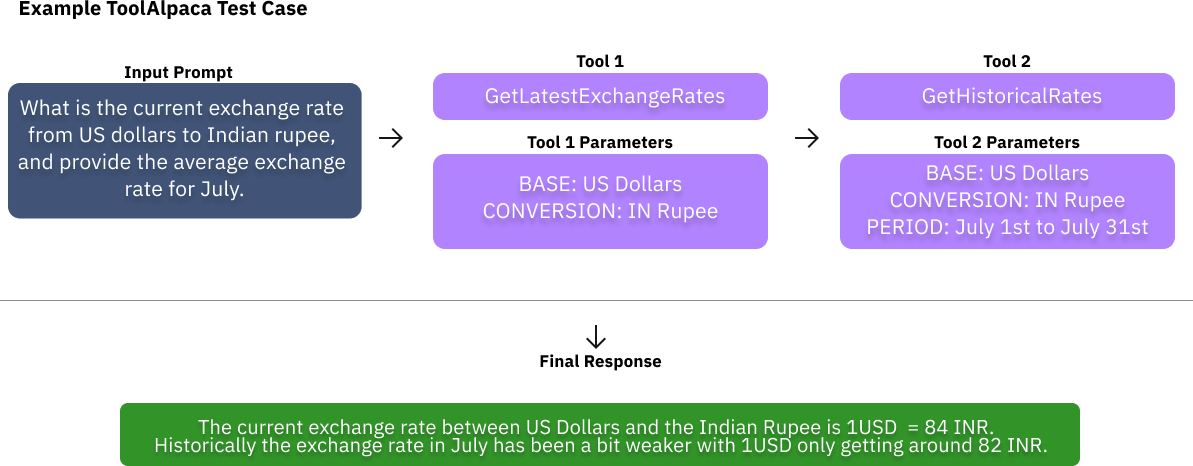}
    \caption{Example test case from the ToolAlpaca test dataset requiring multiple finance domain APIs for task completion.}
    \label{fig:currencycase}
\end{figure}

For evaluating whether a coalition of open-source pretrained (non-fine-tuned) models can be used in the AI agent domain in-place of fine-tuned models, we use the ToolAlpaca test set which contains \(114\) test cases requiring \(11\) real cloud based services. These services and their functionalities are well documented, span a variety of domains (such as geographical, aeronautical, finance and entertainment) and include tests of varying difficulty. An average \(1.25\) tools, (API endpoints) from the services, are required per test case, offering a non-trivial challenge in tool planning. An example test case is shown in Figure \ref{fig:currencycase}.

The developers of this test set also fine-tune the Vicuna-7B and Vicuna-13B models \cite{vicuna2023, tang2023toolalpaca} enabling a direct comparison of these fine-tuned models for tool use versus our coalition of non-fine-tuned models approach.

We utilise the ToolAlpaca \citep{tang2023toolalpaca} test set along with custom built datasets to show that models working together, in a coalition, are more accurate then models working singularly for all tasks. As a result we show that, AI agents utilising the coalition approach can be more robust, by being more accurate, then those reliant on a single model approach.

\subsection{Coalitions of pretrained models outperform fine-tuned models}

\begin{table}
\centering
\begin{tabular}{|c|c|c|c|c|}
    \hline
\textbf{Configuration} & \begin{tabular}[x]{@{}c@{}}\textbf{Planner}\\\textbf{Accuracy}\end{tabular} & \begin{tabular}[x]{@{}c@{}}\textbf{Slot Filling}\\\textbf{Accuracy}\end{tabular} & \begin{tabular}[x]{@{}c@{}}\textbf{Overall}\\\textbf{Procedural}\\\textbf{Accuracy}\end{tabular} & \begin{tabular}[x]{@{}c@{}}\textbf{Response}\\\textbf{Evaluation}\end{tabular} \\
    \hline\hline

\begin{tabular}[x]{@{}c@{}}\textbf{Coalition}\\\textbf{of Open-Source Models}\\\textbf{(with best coalition)}\end{tabular} & 75.7\% & \textbf{61.3}\% & \textbf{58.6}\% & \begin{tabular}[c]{@{}l@{}}Cos: 0.662\\ RLPre: 0.613\\ RLRec: 0.219\\ RLf:  0.286\end{tabular} \\
\hline
\begin{tabular}[x]{@{}c@{}}ToolAlpaca\\(using ToolAlpaca-7b model)\end{tabular} & 82.6\% & 54.1\% & 45.6\% & \begin{tabular}[c]{@{}l@{}}Cos: 0.681\\ RLPre: 0.610\\ RLRec: 0.232 \\ RLf: 0.300\end{tabular}   \\
\hline
\begin{tabular}[x]{@{}c@{}}ToolAlpaca\\(using ToolAlpaca-13b model)\end{tabular} & \textbf{85.4}\% & 57.3\% & 48.2\% & \begin{tabular}[c]{@{}l@{}}Cos: 0.725\\ RLPre: 0.635\\ RLRec: 0.264\\ RLf: 0.342\end{tabular}\\
\hline
\begin{tabular}[x]{@{}c@{}}ToolAlpaca\\(using Vicuna-13b model)\end{tabular} & 64.5\% & 48.4\% & 13.2\% & \begin{tabular}[c]{@{}l@{}} Cos: 0.644\\ RLPre: 0.541\\ RLRec: 0.238\\ RLf: 0.292\end{tabular}\\
\hline
\end{tabular}
\caption{Candidate coalition is assessed against the ToolAlpaca system utilising the Vicuna-13b model and the fine-tuned ToolAlpaca-7B and ToolAlpaca-13B models. The overall procedural accuracy assesses how accurately the system and configuration plan tool usage and execute the tools with the right parameters. The response evaluation column reports on the similarity of the system's final response to human defined responses. The \textit{Cos} metric represents a measure of the semantic textual similarity between the defined responses and the final response. RLPre, RLRec amd RLf measurements report on the similarity based on the RougeL metrics.}
\label{table:fine-tuningvs}
\end{table}

Using the evaluation criteria and metrics detailed in Section \ref{experimentaldetails}, Table \ref{table:fine-tuningvs} shows that a coalition of pretrained open-source models outperforms the fine-tuned ToolAlpaca 7B and 13B models. A system which utilises the coalition of general purpose models achieves an overall accuracy of 58.6\%, beating the fine-tuned models by more than 10\%, without incurring any of the costs associated with fine-tuning. 

The gains in performance can be attributed to the fact that planning, slot filling and response forming tasks have been split in our system and can be assigned to different models, each exhibiting strengths in that specific task. This approach also allows for greater observability of the system enabling programmatic means for capturing and correcting generic hallucinations. An example of this is the ability to capture non-existent tools in the generated plans and removing or replacing those steps with real tools. In contrast, the fine-tuned models are challenged with performing well in each of these tasks independently without opportunities for intervention. We note that the fine-tuned models often fail to meet the requirements of the system or the prompt, triggering parsing errors and downstream failures.

Another major benefit of being able to utilise open-source pretrained models instead of fine-tuning models is that our system is able to plug and play newer, more performant models as they emerge in the community and leverage any improvements in capabilities to increase the accuracy of the overall agentic system.

\subsection{Coalitions outperform using single models}

\begin{table}
\centering
\begin{tabular}{|c|c|c|c|c|}
    \hline
\textbf{Configuration} & \begin{tabular}[x]{@{}c@{}}\textbf{Planner}\\\textbf{Accuracy}\end{tabular} & \begin{tabular}[x]{@{}c@{}}\textbf{Slot Filling}\\\textbf{Accuracy}\end{tabular} & \begin{tabular}[x]{@{}c@{}}\textbf{Overall}\\\textbf{Procedural}\\\textbf{Accuracy}\end{tabular} & \begin{tabular}[x]{@{}c@{}}\textbf{Response}\\\textbf{Evaluation}\end{tabular} \\
    \hline\hline

\begin{tabular}[x]{@{}c@{}}\textbf{Coalition}\\\textbf{Planner: Mistral}\\\textbf{Executor: Mixtral}\\\textbf{JSON RAG: Flan}\\\textbf{Response Former: Mixtral}\end{tabular} & \textbf{75.7}\% & \textbf{61.3}\% & \textbf{58.6}\% & \begin{tabular}[c]{@{}l@{}}\textbf{Cos: 0.662}\\ \textbf{RLPre: 0.613}\\ \textbf{RLRec: 0.219}\\ \textbf{RLf:  0.286}\end{tabular} \\
\hline
\begin{tabular}[x]{@{}c@{}}Coalition\\Planner: Llama 2 70B Chat\\Executor: Mixtral\\JSON RAG: Flan\\Response Former: Mixtral\end{tabular} & 69.3\% & 55.3\% & 49.1\% & \begin{tabular}[c]{@{}l@{}}Cos: 0.652\\ RLPre: 0.585\\ RLRec: 0.182\\RLf: 0.245\end{tabular}    \\
\hline
\begin{tabular}[x]{@{}c@{}}Non-Coalition\\Mistral\\for all LLM\\interactions\end{tabular} & 75.7\% & 56.1\% & 53.5\% & \begin{tabular}[c]{@{}l@{}}Cos: 0.662\\ RLPre: 0.598\\ RLRec: 0.178 \\ RLf: 0.238\end{tabular}   \\
\hline
\begin{tabular}[x]{@{}c@{}}Non-Coalition\\Mixtral\\for all LLM\\interactions\end{tabular} & 70.2\% & 60.5\% & 54.3\% & \begin{tabular}[c]{@{}l@{}}Cos: 0.650\\ RLPre: 0.572\\ RLRec: 0.204 \\ RLf: 0.268\end{tabular}   \\
\hline
\begin{tabular}[x]{@{}c@{}}Non-Coalition\\LLama 2 70B Chat\\for all LLM\\interactions\end{tabular} & 69.3\% & 49.1\% & 43.0\% & \begin{tabular}[c]{@{}l@{}}Cos: 0.676\\ RLPre: 0.625\\ RLRec: 0.144 \\ RLf: 0.209\end{tabular}   \\
\hline
\begin{tabular}[x]{@{}c@{}}Non-Coalition\\Codellama 34B\\for all LLM\\interactions\end{tabular} & 60.5\% & 51.8\% & 45.6\% & \begin{tabular}[c]{@{}l@{}}Cos: 0.612\\ RLPre: 0.520\\ RLRec: 0.231 \\ RLf: 0.058\end{tabular}   \\
\hline
\begin{tabular}[x]{@{}c@{}}Non-Coalition\\Flan\\for all LLM\\interactions\end{tabular} & 29.8\% & 6.1\% & 3.5\% & \begin{tabular}[c]{@{}l@{}}Cos: 0.459\\ RLPre: 0.343\\ RLRec: 0.242 \\ RLf: 0.257\end{tabular}   \\
\hline
\end{tabular}
\caption{Different coalition candidates are assessed against utilising single pretrained models for all tasks. The results identify that models working together for distributed tasks can outperform a single model being assigned all of the tasks. The overall procedural accuracy assesses how accurately the system and configuration plan tool usage and execute the tools with the right parameters. The response evaluation column reports on the similarity of the system's final response to human defined responses. The \textit{Cos} metric represents a measure of the semantic textual similarity between the defined responses and the final response. RLPre, RLRec amd RLf measurements report on the similarity based on the RougeL metrics.}
\label{table:noncoalitionvs}
\end{table}

In Table \ref{table:noncoalitionvs} we show that employing a coalition approach can improve upon using any model singularly. Using the Mistral \cite{jiang2023mistral} and Mixtral \cite{jiang2024mixtral} models independently, for planning, slot filling and response forming the overall system achieves an accuracy of 53.5\% and 54.3\% respectively. When these models work in collaboration with each other, Mistral tasked with the tool planning and Mixtral tasked with slot filling, the overall system accuracy improves to 58.6\%. Similarly, only using the Llama 2 70B Chat model \cite{touvron2023llama} achieves an overall accuracy of 43\%, brought down by its lack of accuracy in the slot filling task. But when Llama 2 is coupled with other models, in a coalition, the system improves its accuracy to 49.1\% since the slot filling task can be offloaded to a model better suited to that task - validating the benefits of using a multi-model approach.

The Mistral, Mixtral and Llama 2 models all show good performance in the planning task. This could be attributed to the data they have been trained on, but could also be equally due to their model architectures.  Each of these models utilitse Grouped Query Attention \cite{ainslie2023groupedqueryattention} which groups tokens and focuses attention to groups containing the most relevant information. This potentially helps the models focus better on important features by removing additional verbosity in the prompts and hence these models may be better equipped to handle planning problems. 

The Mixtral model is built using a Mixture of Experts \cite{jiang2024mixtral, fedus2022mixtureofexpertsreview} architecture and this may be the differentiating factor in its dominant performance at the slot filling task. In this architecture, the tokens are routed to specific experts at each layer. The Mixtral paper references that each of these experts may have developed specialisation to different domains, including specialisation for code, which may enable the model to perform better at slot filling as the model is being tasked with generating correctly formatted request payloads. The Mixtral model used in the coalition is a fine-tuned variant tuned for instruction following, enabling it to accurately respond to the slot filling prompts, following the specified response structure which again allows the system to process the model responses and capture hallucinations more easily, improving the accuracy of the overall system.  

Taking advantage of the best model at planning (Mistral) and the best model at slot filling (Mixtral) we are able to gain a `cost' free accuracy improvement as no additional development, resources or fine-tuning was required to realize this performance upgrade. This offers validation to the idea that allocating tasks to different models unlocks accuracy gains as specific models may be better in different domains. The next section investigates this further.

\subsubsection{Specific models are better specific tasks}

The LLM community has been competing to build the `one model to rule them all'. But in this section we show that certain models exhibit traits, enabling them to perform better at specific tasks and we ask whether utilising multiple models in complex workflows can offer performance improvements beyond using the same, single, model for all tasks.

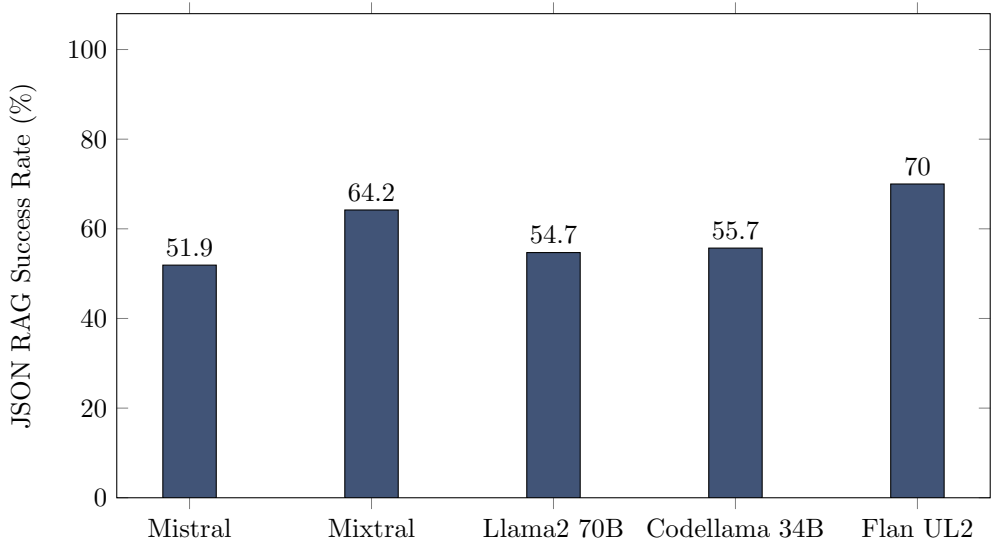
\begin{figure}
\centering
\begin{tikzpicture}
    \begin{axis}[
        ybar,
        ymin=0,
        ymax=90,
        width=\linewidth,
        height=8cm,
        bar width=20pt,
        ylabel={JSON RAG Success Rate (\%)},
        nodes near coords,
        symbolic x coords={Mistral, Mixtral, Llama2 70B, Codellama 34B, Flan UL2},
        xtick = data,
        enlarge y limits={value=0.2,upper},
        legend pos=north west
    ]
    \addplot[fill=darkblue1] coordinates {(Mistral, 51.9) (Mixtral, 64.2) (Llama2 70B, 54.7) (Codellama 34B, 55.7) (Flan UL2, 70.0)};
    \end{axis}
\end{tikzpicture}
\caption{Assessing different models for JSON RAG. The models have been assessed against a custom dataset to determine which model offers the best performance at the JSON RAG task which is applied to filter long JSON responses from tool executions. Flan UL2 20B model outperforms all other model choices at this task.}
\label{fig:jsonragvs}
\end{figure}

Flan UL2 \cite{tay2022ul2} is not the best model by generalised metrics and benchmarks \cite{hendrycks2020mmlu, mmluresults}. But it outperforms all other models considered, at a very specific task we call JSON RAG. A common issue when using LLMs is the restriction of the context windows they can operate on. In the tool use domain, the responses of tools can be extremely large and often exceed these context windows. To overcome this problem, we devised a solution to filter and retrieve the meaningful contents from the JSON responses of the executed tools. This very specific approach is utilised in slot filling tool parameters following previous tool invocations and also by the response former to create a meaningful final response based on the responses of the executed tools. Using the evaluation strategy defined in \ref{critevaluation}, Figure \ref{fig:jsonragvs} shows that the Flan UL2 model outperforms other larger and more capable models in this very particular task of JSON RAG. 

The implementation of the JSON RAG task is very similar to what can be considered a denoising task where the model is asked to fill in masked tokens with tokens already presented in the prompt. An example of the JSON RAG task is demonstrated in Figure \ref{fig:jsonrageval} . Investigating the architecture of the Flan UL2 model we note that the model has been built using a Mixture of Denoisers approach \cite{tay2022ul2}. This potentially offers an explanation as to why this model exhibits considerable performance improvement over the other model candidates in the coalition.

\begin{figure}
\centering
\begin{tikzpicture}
    \begin{axis}[
        ybar,
        ymin=0,
        ymax=90,
        width=\linewidth,
        height=10cm,
        bar width=12pt,
        ylabel={Planner Critique Success Rate (\%)},
        symbolic x coords={Mistral, Mixtral, Llama2 70B, Codellama 34B, Flan UL2},
        xtick = data,
        enlarge y limits={value=0.2,upper},
        legend cell align=left,
    ]
    \addplot[fill=darkblue1] coordinates {(Mistral, 57.7) (Mixtral, 62.9) (Llama2 70B, 68.6) (Codellama 34B, 63.2) (Flan UL2, 9.6)};
    \addplot[fill=darkpink1] coordinates {(Mistral, 52.0) (Mixtral, 70.6) (Llama2 70B, 57.0) (Codellama 34B, 61.6) (Flan UL2, 10.3)};
    \addplot[fill=darkpurple1] coordinates {(Mistral, 67.8) (Mixtral, 86.6) (Llama2 70B, 75.9) (Codellama 34B, 84.2) (Flan UL2, 19.3)};

    \legend{General Critique,Assisted Critique,Explicit Critique}
    \end{axis}
\end{tikzpicture}
\centering
\caption{Assessing plan critiquing capabilities by model and critique explicitness. The models have been assessed against a custom dataset to determine which model offers the best performance at the correction of generated plans and the level of explicitness a model requires to make that correction (General, assisted and explicit critique). Assisted critique refers to critique specific to the category of error such as critique based on ordering or too many steps, explicit critique provides exactly the correction to be made. For example specifying which steps need to be reordered or removed. General critique refers to an error agnostic critique message.}
\label{fig:critiqueexplicitvs}
\end{figure}
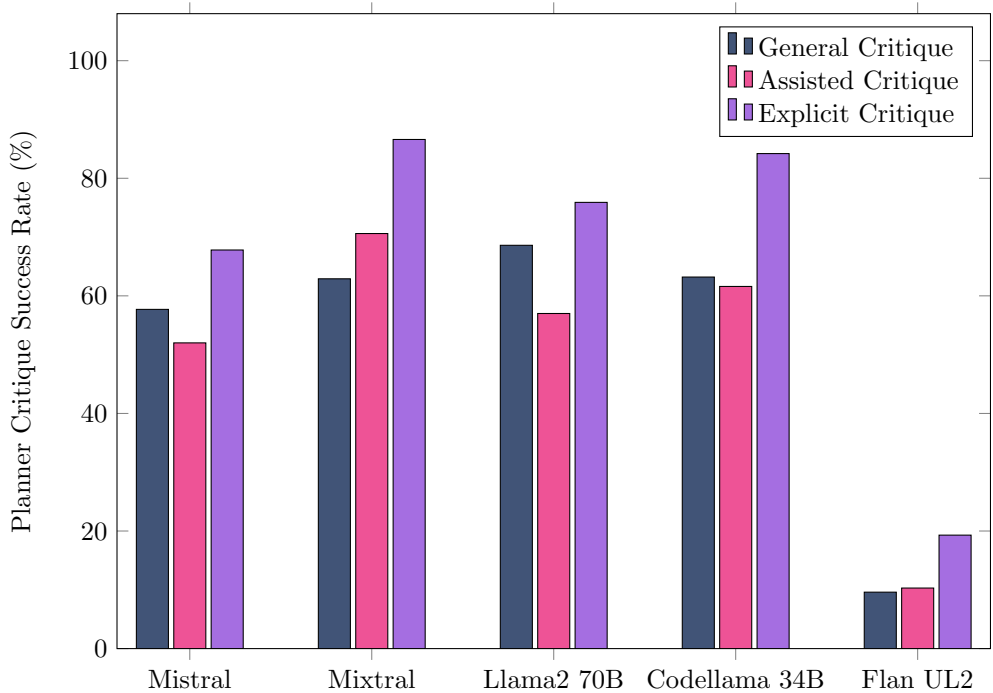

\begin{figure}
\centering
\begin{tikzpicture}
    \begin{axis}[
        ybar,
        ymin=0,
        ymax=90,
        width=\linewidth,
        height=10cm,
        bar width=10pt,
        ylabel={Planner Critique Success Rate  (\%)},
        symbolic x coords={Mistral, Mixtral, Llama2 70B, Codellama 34B, Flan UL2},
        xtick = data,
        enlarge y limits={value=0.3,upper},
        legend cell align=left,
        legend style={font=\small},
    ]
    \addplot[fill=darkblue1] coordinates {(Mistral, 77.2) (Mixtral, 84.5) (Llama2 70B, 88.0) (Codellama 34B, 58.8) (Flan UL2, 17.5)};
    \addplot[fill=darkpink1] coordinates {(Mistral, 44.4) (Mixtral, 58.8) (Llama2 70B, 62.6) (Codellama 34B, 54.7) (Flan UL2, 17.3)};
    \addplot[fill=darkpurple1] coordinates {(Mistral, 60.5) (Mixtral, 74.9) (Llama2 70B, 56.4) (Codellama 34B, 77.8) (Flan UL2, 8.8)};
    \addplot[fill=darkteal1] coordinates {(Mistral, 54.4) (Mixtral, 75.4) (Llama2 70B, 61.7) (Codellama 34B, 87.4) (Flan UL2, 8.8)};

    \legend{Reorder steps,Add required step,Remove step, Remove multiple steps}
    \end{axis}
\end{tikzpicture}
\centering
\caption{Assessing model ability to correct different types of plan failures. The models have been assessed against a custom dataset to determine which model offers the best performance at the correcting generated plans containing different categories of failures such as badly ordered plans, plans missing steps or plans containing too many steps.}
\label{fig:critiquetypesvs}
\end{figure}
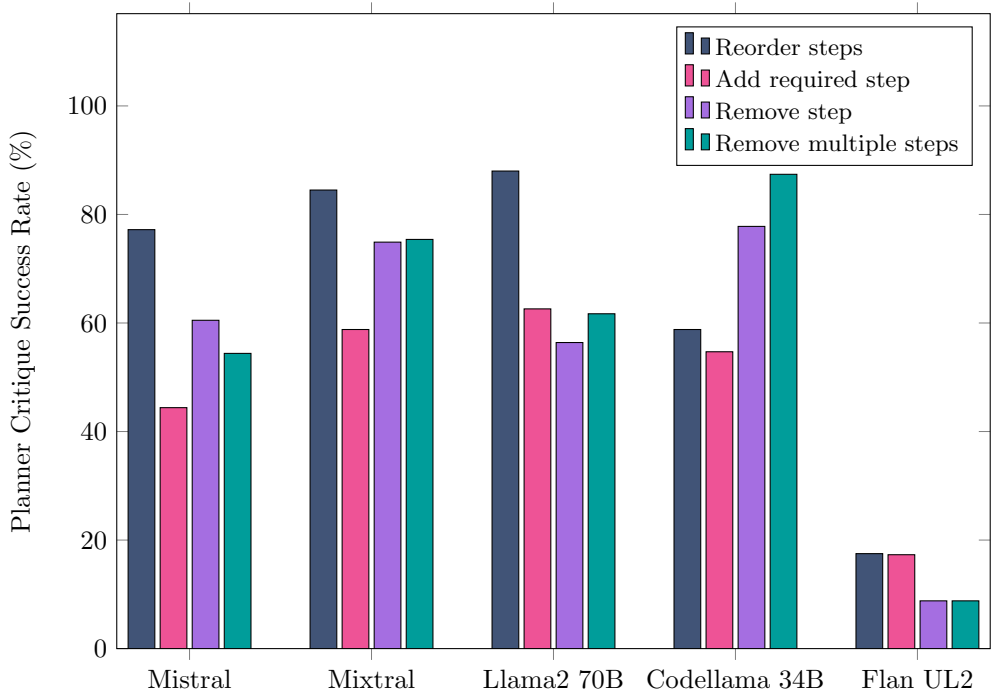

Although Flan UL2 shines in the JSON RAG task, it does not perform well at others. Our coalition system makes use of critique prompting to improve upon the initial generated plans. This involves utilising another model to critique the plan generated by the model used for the planning task in an aim to remove hallucinations and generally improve the plans, an example of critiquing is demonstrated in Figure \ref{fig:criteval}. Using the evalation strategy defined in Section \ref{critevaluation}, Figures \ref{fig:critiqueexplicitvs} and \ref{fig:critiquetypesvs} show that Flan UL2 is easily the least effective model for critiquing and correcting plans. 

For a model to be able to effectively critique complex plans it needs to be able to evoke exemplar reasoning and interactive linguistic capabilities. These capabilities are often acquired through fine-tuning or reinforcement learning with human feedback \cite{reflexion, ouyang2022aligning}, replicating reflection humans perform when assessing a chain of reasoning.  The Mistral, Mixtral, Llama 2 70B and Codellama models assessed in this coalition, have all been fine-tuned for instruction following or for interactive, role-based prompting. These tuning efforts are missing from the Flan UL2 20B model, offering some insights as to why this model is the least accurate when tasked with critiquing generated plans.

Combining the facts that Flan UL2 is the best at JSON RAG and the worst at planner critiquing, we can conclude that the best way to build an accurate system, without requiring tuning, would be to utilise different models for these different tasks - naturally identifying the benefit (and need in this case) for coalition based approaches for agentic systems.

\subsection{Model specialisation leads to accuracy improvements and cost savings}

\begin{figure}
\centering
\begin{tikzpicture}
\begin{axis}[enlargelimits=0.2, ylabel={JSON RAG Success Rate  (\%)}, xlabel={Estimated number of model parameters (Billions)}]
    \addplot[
        scatter/classes={a={darkblue1}, b={darkpink1}},
        scatter, mark=*, only marks, 
        scatter src=explicit symbolic,
        nodes near coords*={\Label},
        visualization depends on={value \thisrow{label} \as \Label} 
    ] table [meta=class] {
        x y class label
        7 51.9 a Mistral
        45 64.2 a Mixtral
        70 54.7 a Llama2
        34 55.7 a Codellama
        20 70.0 b Flan
    };
\end{axis}
\end{tikzpicture}
\caption{Assessing the effectiveness of a model for the JSON RAG task against the model's parameter count. The cost of inferencing scales with model parameter size as higher parameters require more GPU memory for hosting. The Flan UL2 model outperforms all other models for JSON RAG while being a mid sized model.}
\label{fig:jsonragcost}
\end{figure}
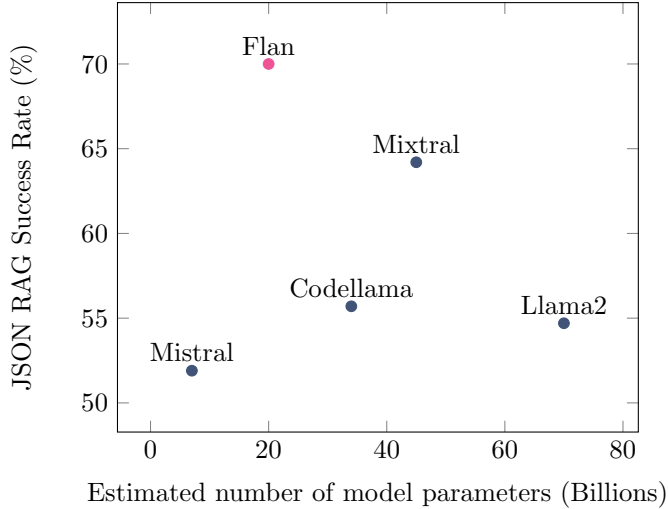

\begin{figure}
\centering
\begin{tikzpicture}
\begin{axis}[enlargelimits=0.2, ylabel={Planner Success Rate (\%)}, xlabel={Estimated number of model parameters (Billions)}]
    \addplot[
        scatter/classes={a={darkblue1}, b={darkpink1}},
        scatter, mark=*, only marks, 
        scatter src=explicit symbolic,
        nodes near coords*={\Label},
        visualization depends on={value \thisrow{label} \as \Label} 
    ] table [meta=class] {
        x y class label
        7 75.7 b Mistral
        45 70.2 a Mixtral
        70 69.3 a Llama2
        34 60.5 a Codellama
        20 29.8 a Flan
    };
\end{axis}
\end{tikzpicture}
\caption{Assessing the effectiveness of a model for the planning task against the model's parameter count. The Mistral model exhibits the best performance and is the smallest of models considered, only using 7 billion parameters, making it the cheapest to use.}
\label{fig:plannercost}
\end{figure}
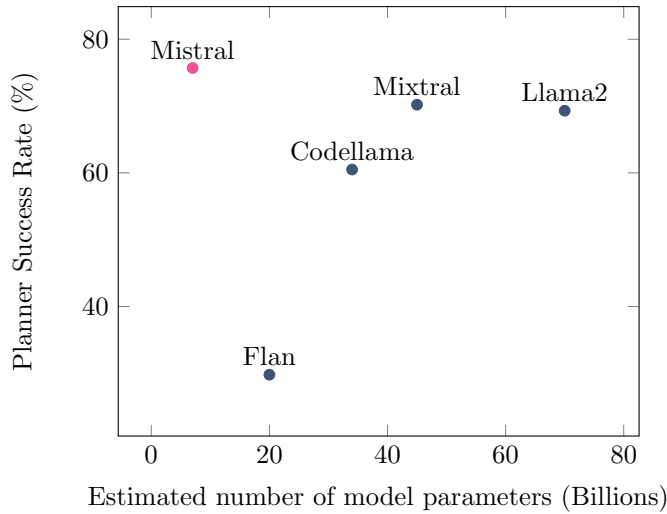

The previous sections have demonstrated that allocating different tasks to different models can improve the performance and accuracy of the overall system. Our results also challenge the trend: `bigger models are better models' \cite{sevilla2022trendsinai}. Although Flan UL2 is a 20B parameter model which is much smaller then the Mixtral (45B), Codellama (34B) and Llama2 (70B) models \cite{tay2022ul2, jiang2024mixtral, roziere2023code, touvron2023llama}, the previous discussion showed that Flan UL2 can outperform these larger models at specific tasks. 

Models with larger parameter counts, impose larger hosting costs since GPU requirements scale with the number of parameters \cite{hfaccelerate} and larger hosting costs are passed onto end consumers through larger inference costs. 

By splitting up tasks and using different models for the individual tasks, a system may be able to use smaller models to gain accuracy improvements while concurrently reducing the cost of the overall system. By recalling the relationship between inference cost with the model parameter count, in Figure \ref{fig:jsonragcost} we see that while Flan UL2 is smaller then most of the other models (hence cheaper), it still outperforms the other models for the JSON RAG task.

Similarly, utilising the coalition approach, when models exhibit similar performance for a specific task, the smallest model can be selected for that task, to cut expenses of the overall system. Considering the choice of model for the tool planning task, in Figure \ref{fig:plannercost}, Mistral, Mixtral and Llama2 show similar accuracy but since Mistral is so much smaller then the other two models, the coalition can utilise Mistral and operate with lower overheads. 

Beyond an academic setting building a system which can benefit from these cost reductions, while maintaining system accuracy, will enable cheaper and faster pathways to productisation and utilisation.

\section{Conclusion}

In this paper we introduced a coalition approach, comprising only of open-source pretrained models, as an alternate methodology to fine-tuning, for LLM powered, tool use agents. This new approach utilises a carefully curated coalition of multiple models, where each sub-task in an agent workflow, is allocated to the best performing model. 

This configuration allows us to choose the right model for the job and proves to be more robust, than using any single model (fine-tuned or not) for all tasks.

Since models are allocated specific tasks, we find that this unlocks the ability to plug and play models from a diverse selection, and utilise smaller models for certain tasks, reducing energy consumption and associated costs without trading off accuracy or performance. 

We evaluate this coalition against the open-source ToolAlpaca dataset, focused on tool use agents, and show that agents using this coalition of open-source pretrained models are more accurate then agents reliant on single fine-tuned models. 

This work suggests that LLM system developers should consider whether using a multi-model approach can offer cost-savings and performance enhancements in their workflows. We conclude with a question for consideration in a future study: \textit{If a coalition of pretrained models can outperform a single fine-tuned model, can a coalition of fine-tuned models achieve state of the art performance?}

\section{Experimental} \label{experimentaldetails}

The ToolAlpaca test dataset was extended and utilised to measure the accuracy of the coalition based approach compared to single model approaches. The different evaluations featured comparing generated plans, tool parameters used and the final responses of the systems. 

The evaluation criteria for these categories was adapted from that used by the ToolAlpaca study to reflect desired behaviour in these agentic systems. 

\subsection{Evaluating Planning}

\begin{figure}
    \centering
    \includegraphics[width=1\linewidth]{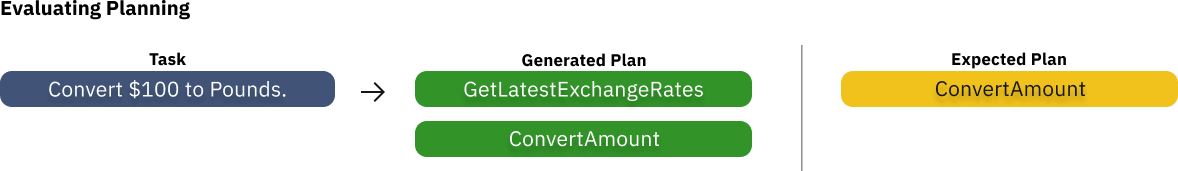}
    \caption{Example of the planning evaluation criteria being met.}
    \label{fig:planeval}
\end{figure}

To evaluate the planning capabilities of the candidate systems and configurations, plans that matched expected, ground truth plans (which we refer to as golden plans), were considered as passes. Additionally, non-optimal plans which contained the golden plans as the final steps were marked as passes. This additional criterion was added as we believe the system should not be penalised for taking additional preliminary steps to perform data gathering and prefer non-optimal plans over those which use fewer steps but assume knowledge based on the pretraining of the model. For an example of this behaviour see Figure \ref{fig:planeval}.

\subsection{Evaluating Slot Filling (Parameter Inference)}

\begin{figure}
    \centering
    \includegraphics[width=1\linewidth]{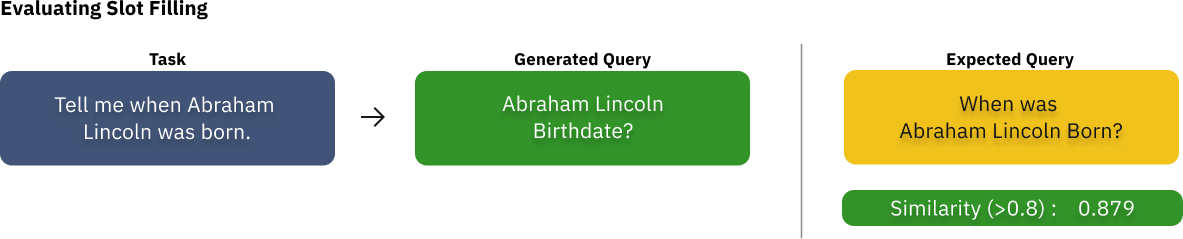}
    \caption{Example of the slot filling evaluation criteria being met in case of query based parameters.}
    \label{fig:sfeval}
\end{figure}

For evaluating the slot filling capabilities of these systems
we measured success based on whether the system inferred the required tool parameters. Since the ToolAlpaca test set makes use of some APIs which require query-based  question answering tools, we also assess whether the generated query was semantically similar to the expected query by measuring the semantic textual similarity and thresholding against a high value. For an example see Figure \ref{fig:sfeval}.

\subsection{Evaluating Procedural Accuracy}

To measure the systems’ procedural accuracy, we measured the number of cases which passed both, the planning and slot filling, evaluations. The overall procedural accuracy assesses how accurately a system and configuration is able to plan for tool use and execute the tools with the right parameters, successfully completing the desired workflow. 

\subsection{Evaluating System Responses}

The response evaluation is used to examine the effectiveness of the final response. Here we report a mixture of Semantic Textual Similarity (STS) and RougeL metrics between the system responses and hand-crafted responses expected to be present in the final response. The STS measurement is performed by using the \textit{all-MiniLM-L6-v2} model to retrieve embeddings for the expected and actual responses. This metric is represented in our results tables as the \textit{Cos} metric. Similarly, the RougeL Precision, Recall and fMearure metrics are represented as \textit{RLPre, RLRec} and \textit{RLf} in our tables respectively. 

Although a LLM-as-a-judge \cite{zheng2024llmjudge} evaluation approach, which utilises a LLM to assess the alignment of a candidate response to the expected response, is an emergent strategy for evaluating system responses, we believe it lacks reproducibility, hence we have favoured the use of the STS and RougeL metrics instead. However, the response metrics should only be considered in partnership with the overall procedural accuracy metric to determine the best performing systems. 

\subsection{Evaluating JSON RAG and Critiquing} \label{critevaluation}

\begin{figure}
    \centering
    \includegraphics[width=1\linewidth]{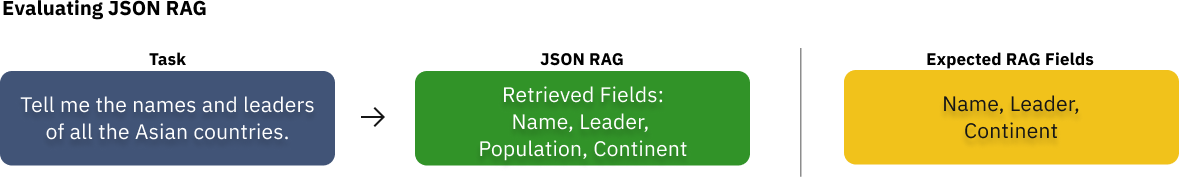}
    \caption{Example of JSON RAG evaluation to select specific contents from a JSON response object from tool execution. In this example, the JSON RAG component is evaluated against its ability to select the country name, country leader and continent based on the given prompt.}
    \label{fig:jsonrageval}
\end{figure}

To evaluate the best model at JSON RAG, a custom dataset was built containing different prompts comprising different JSON contents to be processed. The evaluation assessed whether for a specific prompt/task, the model could select the relevant JSON fields. An example can be seen in Figure \ref{fig:jsonrageval}. The best performing model in this evaluation could then be selected to be used in the coalition.

\begin{figure}
    \centering
    \includegraphics[width=1\linewidth]{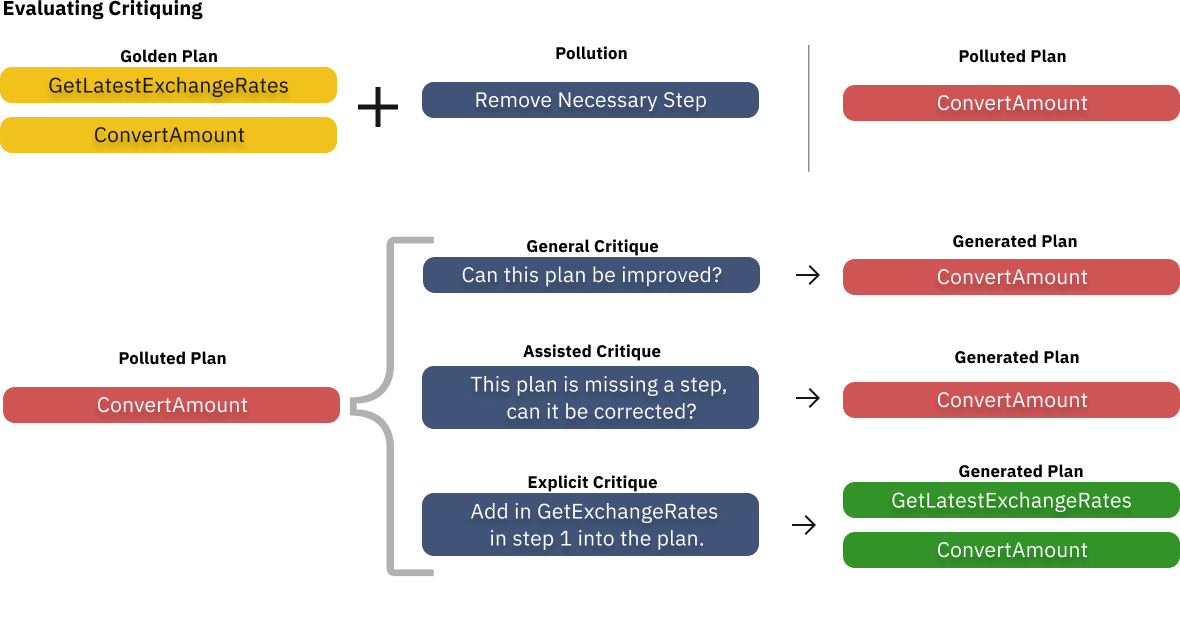}
    \caption{Example of critiquing evaluation with a missing step pollution and tests with varying levels of critiquing feedback to generate the golden plan.}
    \label{fig:criteval}
\end{figure}

To evaluate the best model at plan critiquing, a custom dataset was built containing polluted plans and each model is tasked with correcting the plans based on some feedback. The pollution categories are:
\begin{itemize}
    \item Ordering: Golden plans with the steps reordered.
    \item Missing Step: Golden plans with required steps removed.
    \item Added Step: Golden plans with an unnecessary step added.
    \item Added Multiple Steps: Golden plans with many unnecessary steps added.
\end{itemize}

The model is presented these polluted plans and instructed with feedback to correct the plan. Each model is tested with 3 levels of critique to understand the type of critique they need to correct the pollution:

\begin{itemize}
    \item General Critique - A pollution agnostic, general comment such as \textit{Can this plan be simplified or improved?}.
    \item Assisted Critique - A pollution specific, general comment such as \textit{It seems like the plan has some steps in the wrong order, can the plan be corrected?}.
    \item Explicit Critique - A pollution specific, explicit comment requesting the desired correction such as \textit{Reorder step 1 and 2}.
\end{itemize}

An example of this evaluation can be seen in Figure \ref{fig:criteval}. Given the polluted plans and the various levels of critique, the model which performs the best at correcting the most plans, with the greatest generality of critique can be chosen for the coalition.

\section{Data availability}
Supporting data for this manuscript is available via the link: \href{https://doi.org/10.5281/zenodo.12924336}{https://doi.org/10.5281/zenodo.12924336}.

\section{Code availability}
Implementation to demonstrate the topics covered in this work, as well as generated outputs used for quantitative analysis is available via the link: \href{https://doi.org/10.5281/zenodo.12924336}{https://doi.org/10.5281/zenodo.12924336}. The outcomes of this work can be reproduced using actively developed frameworks for autonomous agent development. 

\section{Acknowledgements}
The authors would like to acknowledge the financial support of the Hartree National Centre for Digital Innovation – a collaboration between the Science and Technology Facilities Council and IBM.

\section{Author contributions}
E.P-K conceived the project and incepted the coalition methodology. P.M. T.K and T.D designed the system, developed the computational pipeline and the evaluation frameworks. P.M, C.M and D.B designed and graded the performance of the system against the open sourced benchmarks, contributing the insights and claims described. P.M, C.M, E.P-K and D.B wrote the manuscript.

\backmatter

\bibliography{sn-bibliography}

\end{document}